\documentclass{article}

\usepackage{arxiv}

\usepackage[utf8]{inputenc} % allow utf-8 input
\usepackage[T1]{fontenc}    % use 8-bit T1 fonts
\usepackage{hyperref}       % hyperlinks
\usepackage{url}            % simple URL typesetting
\usepackage{booktabs}       % professional-quality tables
\usepackage{amsfonts}       % blackboard math symbols
\usepackage{nicefrac}       % compact symbols for 1/2, etc.
\usepackage{microtype}      % microtypography
\usepackage{lipsum}
\usepackage{graphicx}
\usepackage{amsmath}
\usepackage{float}
\title{Perspective Taking in Deep Reinforcement Learning Agents}

\author{
  Aqeel Labash \\
University of Tartu\\
  \texttt{aqeel.labash@gmail.com} \\
   \And
   Jaan Aru\\
   Humboldt University of Berlin\\
   \texttt{jaanaru@gmail.com}\\
      \And
  Tambet Matiisen\\
  University of Tartu\\
  \texttt{tambet.matiisen@ut.ee}
   \And
  Ardi Tampuu \\
  University of Tartu\\
  \texttt{ardi.tampuu@ut.ee}\\
  \And
  Raul Vicente \\
  University of Tartu\\
  \texttt{raulvicente@gmail.com}}
\begin{document}
\maketitle

\begin{abstract}
Perspective taking is the ability to take the point of view of another agent. This skill is not unique to humans as it is also displayed by other animals like chimpanzees. It is an essential ability for social interactions, including efficient cooperation, competition, and communication. Here we present our progress toward building artificial agents with such abilities. We implemented a perspective taking task inspired by experiments done with chimpanzees. We show that agents controlled by artificial neural networks can learn via reinforcement learning to pass simple tests that require perspective taking capabilities. We studied whether this ability is more readily learned by agents with information encoded in allocentric or egocentric form for both their visual perception and motor actions. We believe that, in the long run, building better artificial agents with perspective taking ability can help us develop artificial intelligence that is more human-like and easier to communicate with.
\end{abstract}

% keywords can be removed

\section{Introduction}

Many decisions we take depend on others, what they think, what they believe, and what we know about what they know. This ability to understand and infer the mental states of others is called Theory of Mind \cite{premack1978does} or mindreading \cite{apperly2011mindreaders}. Not only humans have the ability to take into consideration what others think and believe. In controlled experiments it has been shown that chimpanzees can know what other conspecifics see and know \cite{hare2000chimpanzees}. Here we ask whether artificial intelligence (AI) agents controlled by neural networks \cite{Goodfellow-et-al-2016} could also learn to infer what other agents perceive and know.

Theory of Mind is an important topic to study in AI mainly because successful human-machine interaction might critically depend on it. In particular, according to prominent theories, understanding the intentions of others is necessary for the emergence of meaningful communication and language \cite{tomasello2010origins, tomasello2019becoming, scott2014speaking, mercier2017enigma}. According to these views, the basis of communication is not the ability to decode the message from the other, but rather the ability to understand that the other is \textit{trying to communicate} in the first place. Humans excel at language and communication because we have a bias to assume that others are trying to “get something across” to us \cite{tomasello2010origins, tomasello2019becoming, scott2014speaking}. Unfortunately, we do not know how this bias is implemented in the human brain and hence we do not know how to implement it in AI. Furthermore, developing agents that are at the level of humans in Theory of Mind will be much more complicated than in computer or board games, as there are no established tasks or benchmarks to train the agents on.

Under these circumstances one way to proceed is to study how much of the Theory of Mind abilities could be learned through reinforcement learning (RL) in simple tasks. This approach would demonstrate the limits of simple model-free algorithms for understanding Theory of Mind and thus build the basis for more comprehensive approaches. RL is a branch of AI that allows an agent to learn by trial and error while interacting with the environment. In particular, the agent must learn to select the best action in each specific state to maximize its cumulative future reward \cite{sutton2018reinforcement}. The agent could be for example an autonomous robot \cite{DBLP:journals/corr/LevineFDA15} or a character in a video game \cite{mnih2015human}. The idea behind learning by interacting with an environment is inspired from how human and animal infants learn from the rich cause-effect or action-consequence structure of the world \cite{sutton2018reinforcement,thorndike1911animal,schultz1997neural}. Therefore, RL is a biologically plausible mechanism for learning certain associations and behaviors and it can be used to study Theory of Mind.

Theory of Mind was studied by Rabinowitz and colleagues who modelled agents' behavior in a grid world \cite{rabinowitz2018machine}. The proposed neural network was trained using meta-learning; in a first stage the network was trained to learn some priors for the behavior of different types of agents to subsequently speed up the learning of a model of a specific agent from a small number of behavioral observations. Their approach was a first step to induce theory of mind faculties in AI agents that were indeed able to pass some relevant tests for Theory of Mind skills. However, as the authors themselves note, their approach was limited in several important aspects that require future work. To name a few, the observer agent learning to model the behavior of others was trained in a supervised manner, it had full observability of the environment and of the other agents, and it was not itself a behaving agent. 

In the current work we are interested in the emergence of certain aspects of Theory of Mind in behaving agents trained via RL with partial observability. We believe that these are more plausible conditions to model how humans and other animals might develop these abilities. In particular, we test here the ability of agents trained via RL to acquire one essential part of Theory of Mind: \textit{perspective taking}.

\subsection{Perspective Taking}

Perspective taking is the ability to look at things from a perspective that differs from our own \cite{ryskin2015perspective}. It could be defined as "the cognitive capacity to consider the world from another individual's viewpoint" \cite{davis1983measuring}. It is one of the social competencies that underlies social understanding in many contexts \cite{galinsky2008pays, apperly2011mindreaders}. The perspective taking ability is not unique to humans and has been observed in other animals like chimpanzees \cite{hare2000chimpanzees, tomasello2003chimpanzees}. 

Chimpanzee social status is organized hierarchically (dominant, subordinate) \cite{goldberg1997genetic}, which is at full display during food gathering: when there is food available that both can reach, the dominant animal almost always obtains it. But what happens if the dominant could potentially reach the food placed behind an obstacle, but does not know that food is there? Can the subordinate take advantage of this? In a series of experiments \cite{hare2000chimpanzees} two chimpanzees were set into two separate cages facing each other with food positioned between them. The researchers manipulated what the dominant and the subordinate apes could see. For example in one condition, one piece of food could not be seen by the dominant chimpanzee. The results demonstrated that the subordinate animal exploited this favourable situation and indeed obtained more food in this condition. Hence, it was able to consider what the dominant chimpanzee could and could not see, i.e. take the perspective of the dominant chimpanzee into account \cite{hare2000chimpanzees, tomasello2003chimpanzees}. This work done with chimpanzees was the inspiration for our study. 

The aim of the present work is to study whether an AI agent controlled by a neural network can learn to solve a similar perspective taking task using RL. We chose this task because it is relatively simple, while allowing us to study perspective taking with RL. We do not claim that RL captures all aspects of perspective taking or is the exact model of how perspective taking is learned in biological organisms \cite{aru2018deep}. We even do not claim that RL allows us to understand or model how chimpanzees solve this particular task. We simply use this task as to probe whether some simple aspects of perspective taking can be learned by RL agents. Understanding the capabilities and limitations of RL in acquiring perspective taking skills will lead to a better algorithmic understanding of the computational steps required for perspective taking in biological organisms.

\begin{figure}
\centering
\includegraphics[scale=0.73]{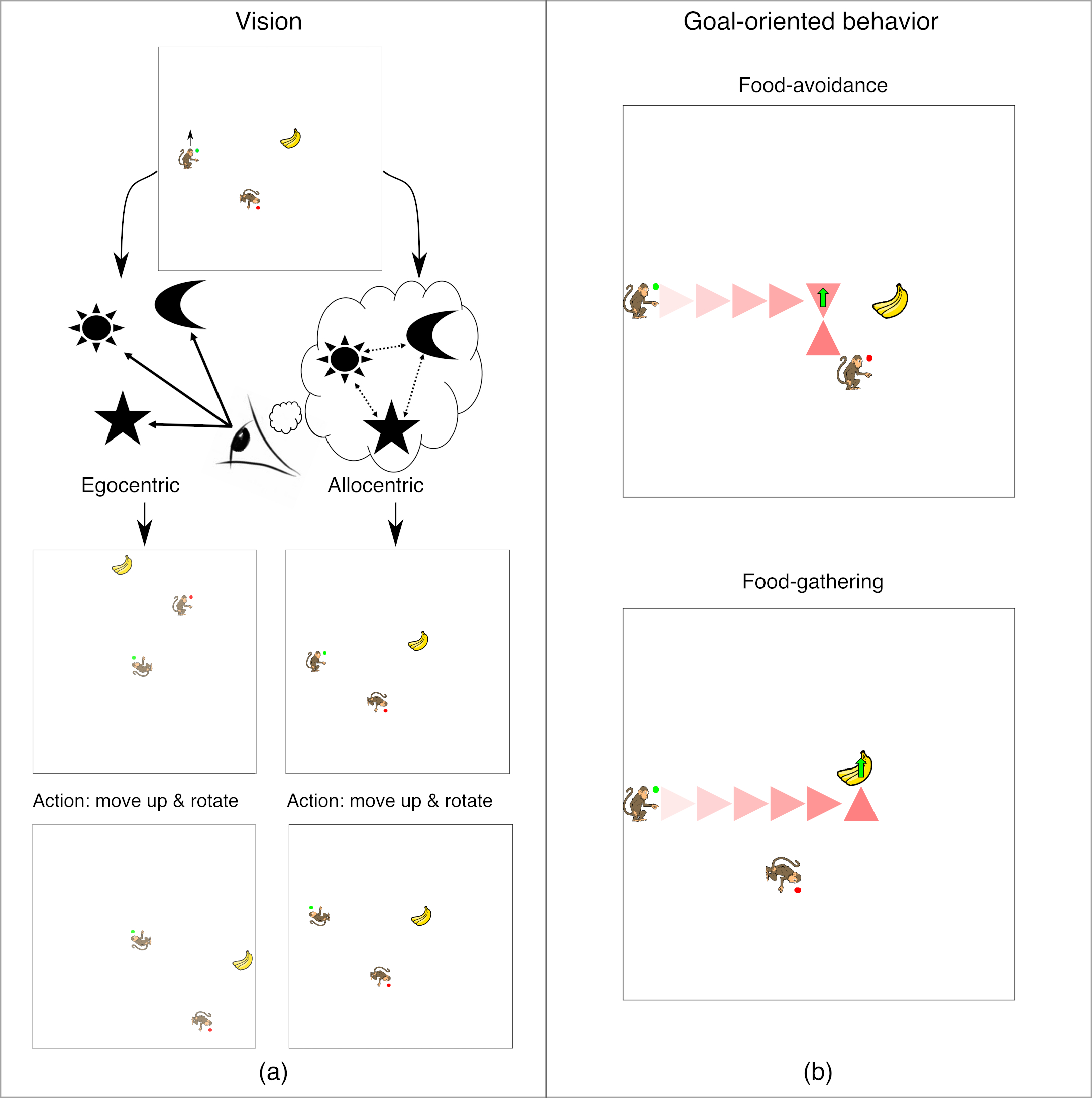}
\caption{\textbf{(a)} Overview of the simulation environment and visual encodings. The artificial monkey with green circle is the subordinate agent and the one with red circle is the dominant agent. The hands of both agents point to their orientation. Below in the same diagram it is illustrated how egocentric and allocentric visual representations differ. In the egocentric mode objects are perceived relative to the perceiving agent's position and orientation. In the allocentric mode the agent perceives the objects in terms of their location with respect to a fixed reference. \textbf{(b)} Two examples of a subordinate agent goal-oriented behavior as driven by our neural network controller. In the top panel the agent should avoid the food as it is observed by the dominant. In the bottom panel the agent should acquire the food as it is not observed by the dominant. The path the agent followed is marked with the red triangles. The green arrow represents the final position and orientation of the subordinate (monkey with green circle) when the episode terminated.}
\label{Fig1}
\end{figure}

Furthermore, we are interested in a specific question about perspective taking: is it simpler to learn perspective taking with allocentric or egocentric representations of the environment? With allocentric input the position of other objects and agents is presented in relation to each other independently of the position of the perceiving agent. With egocentric input the position of all objects and other agents is given with respect to the position of the perceiving agent. This means that for example when the agent changes its orientation the whole world will rotate. See Fig \ref{Fig1} for an illustration of the two encodings of visual input. From neuroscience and behavioral experiments it is known that although animals perceive the world from the egocentric viewpoint, this information is transformed to allocentric code in structures like the hippocampus \cite{burgess2001memory, wilber2014interaction, chersi2015cognitive, wang2020egocentric}. Presumably the fact that this transformation is computed in the brain hints that the allocentric view enables some functions that cannot be achieved through egocentric representation alone \cite{burgess2001memory, chersi2015cognitive}. It is possible that perspective taking is one of these functions. Intuitively it seems that taking the perspective of the other agent demands ignoring own sensory input and taking into account the relations between the other agent and the objects in the environment, which could be supported by allocentric representation. Hence, another goal of our study is to test the generality of these assumptions and intuitions using minimal models of computational learning agents.

\section{Methods}
\subsection{Task and Environment}
For the perspective taking task, we generated a %$11\times11$ 
grid world environment where each element can spawn randomly within specific regions. The elements considered included two agents (a \textit{Dominant} and a \textit{Subordinate}), and a single food item (reward). In the present experiments only the subordinate agent is controlled by a RL algorithm and can execute actions in the environment by moving around and changing its orientation. The dominant agent is not controlled by any learning algorithm but its role is critical. The value of the reward obtained by the \textit{Subordinate} at reaching the food depends on whether the food is visible from the \textit{Dominant}'s point of view. If food is retrieved by the subordinate when observed by \textit{Dominant} the value of the food item becomes negative (to mimic the likely punishment received from the dominant in the nature). If the food is obtained while not observed by the \textit{Dominant} the value of the reward is positive. See Table \ref{tab:reward} for the list the events rewarded and its correspondent values.

\begin{table}[h]
    \centering
    \begin{tabular}{|c|c|}
    \hline
         Event & Reward Value \\
         \hline
         Eating food observed by dominant & -1000 \\
         \hline
         Eating food not observed by dominant & +1000\\
         \hline
         Every time step & -0.1\\
         \hline

    \end{tabular}
    \caption{List of the events rewarded in the experiments and their respective values.}
    \label{tab:reward}
\end{table}

Experiments were conducted using environments created with Python toolbox Artificial Primate Environment Simulator (\textit{APES}) \cite{DBLP:journals/corr/abs-1808-10692}. The toolbox allows to simulate a 2D grid world in which multiple agents can move and interact with different items. Agents obtain information from their environment according to a visual perception model. Importantly, \textit{APES} includes different visual perception algorithms that allows to calculate visual input based on agents' location, orientation, visual range, visual field angle, and obstacles. In particular, for this work, we simulate two modes for the agents' vision: egocentric and allocentric (for detailed descriptions see the subsection on visual encoding). For further specifics on the toolbox the reader can access the associated GitHub repository \footnote{ \href{https://github.com/aqeel13932/APES}{https://github.com/aqeel13932/APES}}.

\begin{figure}[!ht]
%\hspace{-1.1cm}
%\begin{minipage}[r]{1.1\textwidth}
\centering
\includegraphics[width=0.5\linewidth]{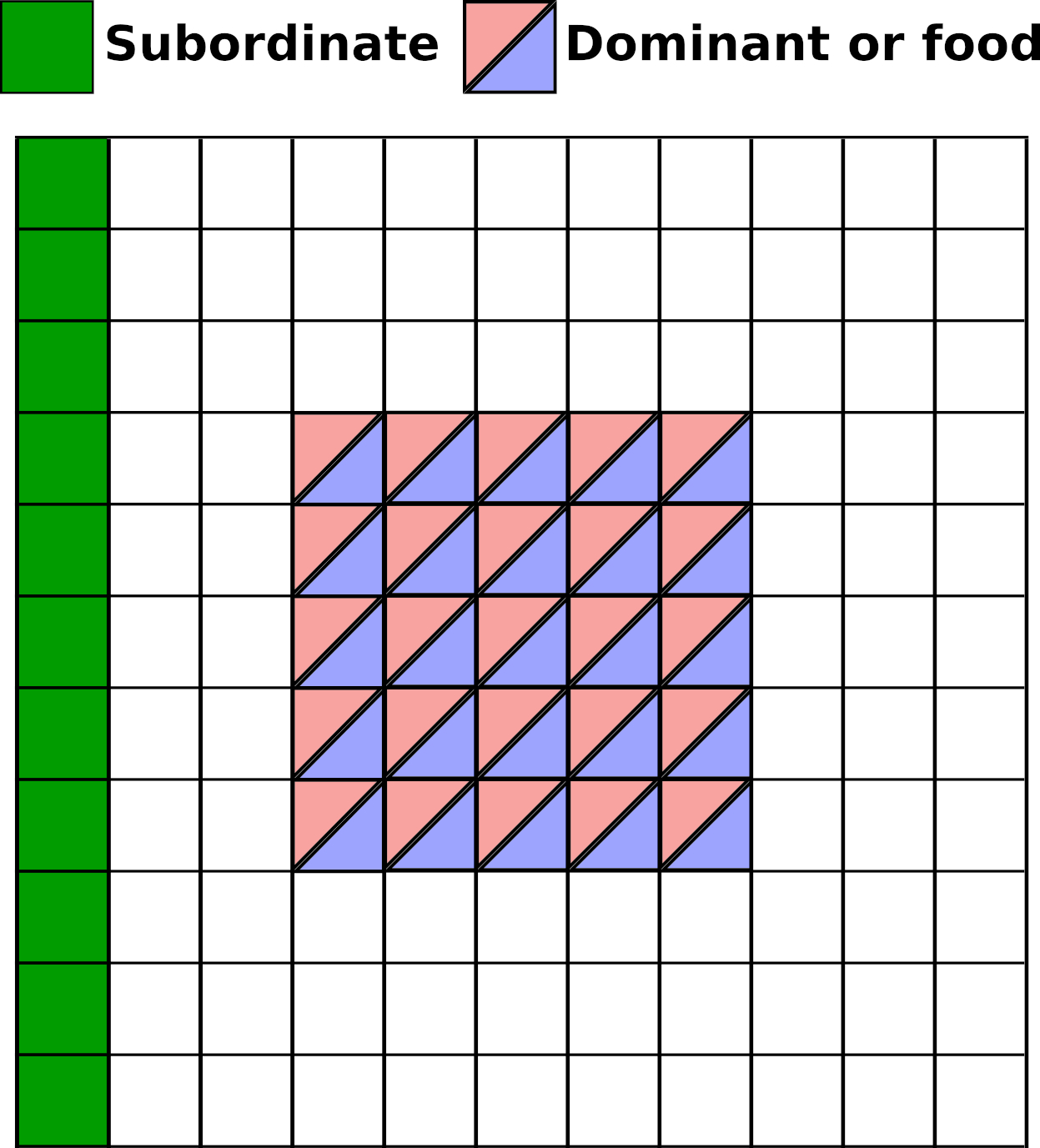}
\caption{Possible starting positions for each element in the environment. The subordinate agent (green) is always spawn in the leftmost column and facing East (looking right) at the start of the episode. %The dominant agent (red) and food (blue) can spawn in larger areas with higher levels of the task. \textbf{a)} In Level 1 the dominant has a fixed position but it has different orientations between episodes. The food has a fixed position. \textbf{b)} In Level 2 the food can spawn in a $5 \times5$ area. \textbf{c)} In Level 3 
The dominant and the food can both spawn anywhere in a $5\times5$ area. Note that overlap between elements is not allowed (food and dominant cannot occupy the exact same cell).}
\label{Fig2}
%\end{minipage}
\end{figure}

In all the experiments we considered that both agents have a long range of vision but a limited visual field angle of 180 degrees. In our main scenario for our perspective taking task, the coverage of the food and dominant's location is distributed as shown in Fig \ref{Fig2}. Both the dominant and food item can spawn anywhere inside a $5 \times 5$ area (see Fig \ref{Fig2}). To successfully solve the task the subordinate agent must learn to navigate to reach the food's location only when the food item is not withing the field of vision of the dominant agent. This implies that the subordinate needs to simultaneously integrate three pieces of information to successfully determine whether the food item is observed by the dominant or not: 1) the orientation of dominant, 2) the position of the dominant, and 3) the position of the food. Note that since the subordinate agent moves and rotates around the environment a direct perception of the dominant agent and food is not always present. However, as explained below the agent is equipped with a short-term memory in the form of a LSTM layer that allows it to integrate temporal information \cite{hochreiter1997long}.

The dimensions of the grid world amounts to $13\times13$ when using allocentric encoding of visual information, and $11\times11$ when using the egocentric encoding. This compensation is needed to balance the fact that egocentric encoding needs a larger input space (since positions are relative to the agent's location and orientation, a $n \times n$ grid world actually needs a $n \times 2n-1$ input layer for an egocentric agent with 180 degrees vision). The dimensions of the grid worlds were chosen so that the neural network controllers for egocentric and allocentric encodings match in their number of parameters (weights). We also matched the average distance between the initial location of the subordinate agent and the food item. 

The number of possible combinations for initial configurations of the environment and agents exceeds the $20000$. Upon movement of the subordinate agent along the grid the number of possible states becomes $>1000000$.

\subsection{Model}\label{model}
\subsubsection{Input}\label{network.input}
The input to the network controlling the subordinate actions is a set of binary maps. They encode the different agents and other elements properties in the environment. The list of inputs to the network include: 

\begin{itemize}
    \item \textbf{Spatial location of elements:} $13\times 13$ or $11\times 21$ binary one-hot map for each element represented by a $1$ at the corresponding element position. In egocentric vision, the agent own location is not required. This is because the focal agent location does not change (in an egocentric framework the agent is always at the center of its own visual field).
    \item \textbf{Observability mask:} $13 \times 13$ or $11 \times 21$ binary mask which indicates the field of vision of the subordinate. It helps to distinguish whether a cell in the grid world is empty or out of the field of vision.   
    \item \textbf{Orientations:} $1 \times 4$ binary one-hot vector for each agent with $1$ at the corresponding agent orientation. In egocentric vision, only the orientation for dominant agent is required since in egocentric the subordinate is always looking forward. An important difference between allocentric and egocentric vision is orientation encoding. In allocentric vision the orientation is encoded as (North, South, East, West) in comparison to egocentric vision where orientation is in relation to the focal agent (towards the agent, same direction as the agent, to its left, to its right).
\end{itemize}

%Note that when using egocentric vision the location maps are $21 \times 21$ to take into account the movement of the agent in the environment.
%\begin{list}

%\end{list}

\subsubsection{Visual encoding: allocentric vs egocentric}
In this work we compare two types of visual perception. With allocentric input the locations and orientations of items in the environment are encoded in reference to a fixed system of coordinates (as if the vision is provided by a fixed camera with a top-down view). With egocentric input, the items are perceived from the eyes of the subordinate agent, and hence they change in relation to the agent movements and rotations. In both allocentric and egocentric there is an observability binary array which represents the observed and non-observed areas.

%where the network perceives the grid-world top-down as if a camera looking to the world from top. And \textbf{Egocentric}, where the network sees the world from the eyes of the subordinate. 

In allocentric encoding, we feed to the network $13 \times 13$ arrays that represent positional information of items in the environment in addition to $1\times 4$ array for each agent in order to encode their orientation. In this mode, when the subordinate changes its orientation and moves, four bits will change corresponding to its previous location, current location, previous orientation, and current orientation. We note that the observability and the dominant agent orientation array might also change in case the subordinate movement leads to some item switch from observed to not observed (or viceversa).

%when the subordinate changes its orientation and moves, it only changes 1 bit in its orientation array, another in its position array, observability array (depending on position and orientation), dominant orientation array in case became observed or not.  %  In this mode, when the subordinate changes its orientation and moves, only four bits will change corresponding to its previous location, current location, previous orientation, and current orientation.

In egocentric encoding, \textit{Subordinate}'s position and orientation remain fixed despite the agent's movements or rotations. We humans, similarly to other animals, when we turn left or right we still look forward and in the same position from our perspective. Hence, in egocentric encoding the network is not fed \textit{Subordinate}'s orientation, but still it is fed the relative \textit{Dominant}'s orientation. Thus, \textit{Dominant}'s orientation input is based on where it looks in relation to the \textit{Subordinate}'s (toward the subordinate, same direction as the subordinate, looking to its left or right). In the egocentric condition the input arrays that represent the environment have dimension $11 \times 21$. Although the base environment is $11 \times 11$, the input layer is augmented in the egocentric encoding to accommodate that the agent (with 180 degrees of vision) should always perceive the environment from in reference to its own centered view and of the same size regardless of its position. For example, when the agent is located at bottom right corner of the environment and looking North (forward from its perspective) it should have the left of its visual field encoding the $11 \times 11$ environment. However, when sitting on the bottom left corner and facing the same orientation, now the environment should be displayed as its right visual field. Hence, the input layer is augmented to $11 \times 21$ to accommodate a common range of centered vision regardless of the agent's location.

%In other words, in egocentric vision things are located in the same vision field i.e whatever you have on your right can be occupied by anything and lacking objects or having a wall does not mean losing that field of vision but it is just unoccupied. This means when our agent at bottom right corner looking North (forward from its perspective) it will have $11\times11$ occupied by environment and rest $11\times10$occupied by nothing. %since the agent can look at opposite orientations from opposite corners of the arena.

\subsubsection{Action encoding: allocentric vs egocentric}

The action space of the network controller (which architecture we explain next) depends on the framework for motor output simulated for the agent. In the allocentric encoding of motor output the action space is composed by (moving North, moving South, moving East, moving West, and no move). In the egocentric encoding of motor output the action space is (move forward, move backward, move right, move left, and no move). Note that each moving action is accompanied by a rotation so that the agent is always looking at the direction is heading. For example, if the agent moves North, it will also rotate to face North. This conforms to the fact that humans and most animals advance in the same direction they are facing. We also note that in most classical video games the combination of allocentric actions and allocentric vision is used.

%In this work, we used two types of actions: egocentric actions and allocentric actions. Egocentric actions consist of moving(forward,backward,right,left,standstill). The egocentric actions are the more natural way to move under egocentric vision as humans use them. Allocentric actions consist of moving (North, South, East, West, standstill). The allocentric actions usually used with allocentric vision and mostly used in games.

\subsubsection{Architecture}
\begin{figure}[!ht]
\begin{center}
\includegraphics[width=\linewidth,angle=-90,origin=c]{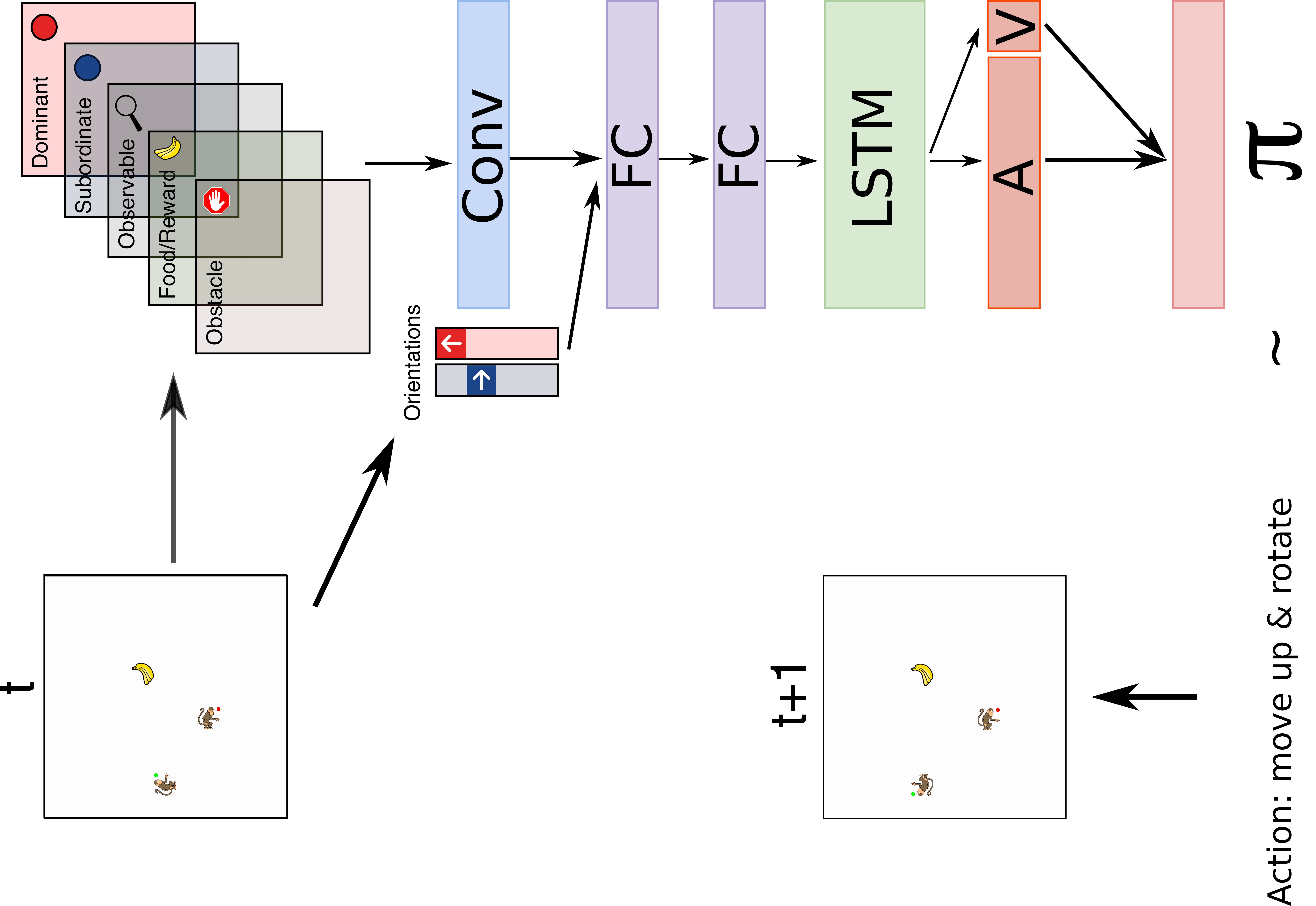}
\caption{The architecture used in the model has 1 convolutional layer with 6 filters and kernel of size 3 followed by 2 fully connected layers with 32 hidden nodes each, and 1 LSTM layer with 128 cells. Output of the LSTM layer is used to learn the  advantages for each possible action $A$ and the state value $V$. Together the state value and advantage heads are used to compute the Q values using Equation \ref{eq:model.mod}, which determines the policy $\pi$ of the agent. Input to the network includes 4 binary matrices that represent: dominant agent, subordinate agent, food item, observable (which is a map that contain 1 at positions of the grid within the field of vision of the subordinate agent and 0 otherwise, on all other maps only observable positions are populated). Each binary matrix indicates the position of an observed element by 1 and zeros otherwise. The orientation of the \textit{Dominant} and \textit{Subordinate} agents is represented by one hot vector of size four for each agent. These vectors are concatenated with the flattened output of the convolutional layer.}
\label{Fig3}
\end{center}
\end{figure}

In our model we used a neural network to control the actions of the subordinate agent. The architecture and hyperparameters are the same as in \cite{DBLP:journals/corr/abs-1810-08647} with two important exceptions.
First, additional inputs are fed to the network. Orientation of both agents are fed after the convolutional layer as shown in Fig \ref{Fig3}. Note that in egocentric encoding only the \textit{Dominant}'s orientation was fed. Second, we used a dueling Q-network \cite{DBLP:journals/corr/WangFL15} instead of the advantage actor-critic model.

%\begin{itemize}
%    \item Additional inputs are fed to the network. In the case of other agent's orientation they are input after the convolutional layer as shown in Fig \ref{Fig3}.
%    \item The original model in \cite{DBLP:journals/corr/abs-1810-08647} produces a state value (which estimates how good the given state is) and policy while the present network produces a state value and advantages like \cite{DBLP:journals/corr/WangFL15}, which  estimate the benefit that different actions bring in the given state.
%\end{itemize}

In a dueling Q-network the state-action value (Q-value) calculation is based on two separate estimates: the state value (how good the current state is) and advantages (which benefit is obtained from each action) as described by %Equation \ref{eq:model.mod} 

\begin{equation} \label{eq:model.mod}
    Q(s,a; \theta) = V(s; \theta_V) +\left[ A(s,a; \theta_A) - \max_{a'} A(s,a'; \theta_A) \right] ,
  \end{equation}

\noindent where $s$ is a state, $a$ is an action, $V$ is the state value, $A$ is the advantage, and $a'$ is the next action. $\theta$ represents the network parameters, while $\theta_V$ is the subset of parameters used in the value network and $\theta_A$ is the subset of parameters used in the advantage network.

Using a dueling network architecture involves updating two network models: a training model (parametrized by $\theta$) which weights are updated using gradient descent, and a target model (parametrized by $\theta^{-}$) which weights are periodically $\tau$-averaged with training model's weights as described by %Equation \ref{target.model}:

\begin{equation}
    \theta^{-} = \theta \tau + \theta^{-} (1-\tau) .
    \label{target.model}
\end{equation}

The $\epsilon$-greedy policy $\pi(s;\theta)$ chooses a random action with probability $\epsilon$ and an action with maximum Q-value otherwise as in Equation \ref{eq:policy.mod}:

\begin{equation} \label{eq:policy.mod}
    \pi(s; \theta) = 
    \begin{cases}
         \text{random action} & u < \epsilon, u \sim U(0, 1) \\
         \mathrm{argmax}_a Q(s, a; \theta) & \text{otherwise}
    \end{cases}
  \end{equation}

%includes moving north, south, west, east, and "not move" for allocentric. And forward, backward, right,left, and "not move" for egocentric. Every moving action is accompanied by a rotation so that the agent is always looking at the same direction it is heading. For example, if the agent moves North, it will also rotate to face North. This conforms to the fact that humans and most animals advance in the same direction they are facing.

To summarize, Fig \ref{Fig3} illustrates the network architecture, its input layers and the output to control the actions of the subordinate agent.

\subsection{Training}
 %for Level 1 and Level 2 were trained for 6 millions steps. Models for Level 3, with highest complexity, 
\subsubsection{Reinforcement models}
All RL models were trained for 20 million steps. An episode is terminated when the food is eaten by the subordinate agent or the food is not eaten after 100 time steps.

%All three levels have one shared reward strategy as explained in Table \ref{tab:reward}. %regardless if they were allocentric or egocentric. 

%Each episode can be up to 100 time steps if the food item was not obtained earlier.

We used replay memory to remove sequential correlations and smooth distribution changes during training as it is usually done in other studies \cite{hausknecht2015deep}. The replay buffer size used is $10^3$ trajectories. Maximum length of each trajectory is 100 time steps. Shorter trajectories were padded with zeros. 

For the neural network implementation we used the \texttt{Keras} library \cite{chollet2015keras}. We used Adam optimizer (batch size set to 16) with a fixed learning rate at $0.001$ and annealed the exploration probability with a schedule from an initial value of $1$ until reaching $0.1$ at the $75\%$ of the total number of steps. We clipped the gradient at 2 to prevent the gradient explosion problem.

\subsubsection{Supervised models}\label{sec:supervized_training}
For the supervised models we used the same network as described in Fig \ref{Fig3} with the exception that
we removed the LSTM layer since we trained the model to predict from the initial time step and hence no memory effects were necessary.

The learning signal was the ground truth label of whether the dominant agent observes the food item within its field of vision for each environment initialization. Thus, the datasets consisted of all possible variations for the initial configurations of the environments (amounting to 26400 samples for egocentric vision and 32100 for allocentric vision). We used a 80/20 split for training and validation data. The training proceeded by minimising the cross-entropy loss function using Adam optimizer (batch size was set to 64 while the learning rate was set to 0.001). To ensure the representativity of the models, all accuracies were averaged over 20 random initializations of the model weights.

%$26400,32100$ samples for egocentric, allocentric respectively and were split 80$\%$-20$\%$ for training and validation. To ensure the representativity of the models, all accuracies were averaged over 20 random initializations of the model weights. 

%Overfitting was not observed as training and testing errors were almost identical.

%2) the output of the network is a single output; 3) the training was supervised not using RL. 

%To test the performance of predicting if the dominant can or cannot see the food from the \textit{Subordinate} input, we used the same network 

\section{Results}

\begin{figure}[!ht]
 \includegraphics[width=\linewidth]{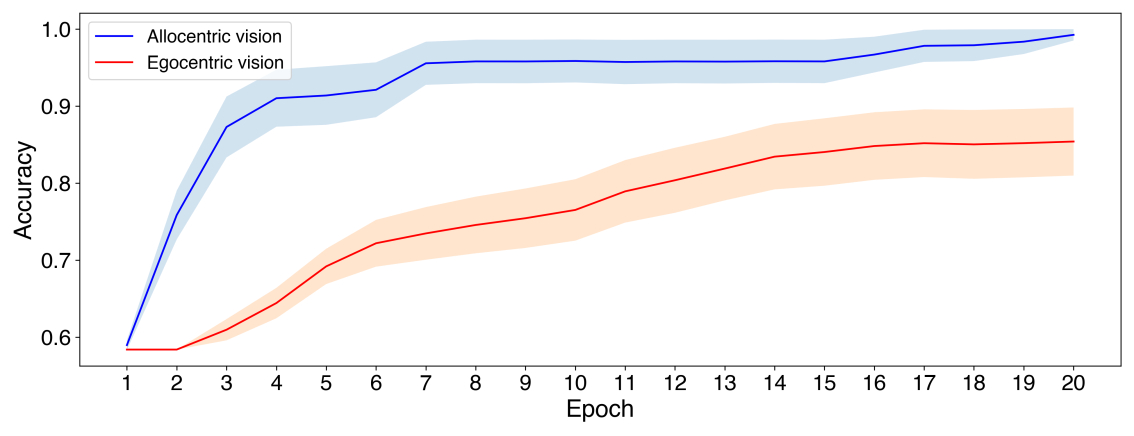}
\caption{Validation accuracy when using allocentric vs egocentric visual inputs to predict whether the food item is observed by the dominant agent or not. In this case the navigational aspects of the task are eliminated, and simple supervised learning was used to train the model. Solid lines and shading indicate the average validation accuracy and standard error of the mean (SEM) over 20 different initializations of the network weights.}
\label{Fig4}

\end{figure}

Here we present the main experiments to test the ability of the agent to solve the present perspective taking task. In particular, we are interested in comparing how the visual and action encoding affects the readiness to learn the task by trial and error. 

As described in Methods, both the dominant agent and food item can spawn anywhere inside a $5 \times 5$ area (see Fig \ref{Fig2}), giving rise to a large number of possible combinations for their relative position and orientation. To solve the task the subordinate agent must learn to navigate to reach the food's location only when the food item is not withing the field of vision of the dominant agent. This implies that the subordinate needs to integrate three pieces of information to successfully determine whether the food item is observed by the dominant or not: 1) the orientation of dominant, 2) the position of the dominant, and 3) the position of the food. In addition, it needs to maintain in memory the result of that integration of information, and navigate successfully towards the food item.   

%Note that since the subordinate agent moves and rotates around the environment a direct perception of the dominant agent and food is not always present.

Thus, solving the perspective taking task presumably involves both an estimation of whether the food item is being observed by the dominant as well as memory and navigational aspects to reach the food item (or avoid it). 

To compare both types of visual processing (egocentric vs allocentric encoding) independently of memory and navigational aspects, first we trained the model to output a binary decision about whether the food item is visible to the dominant in the first time step (initial configuration of each episode). The model architecture is the same as depicted in Fig \ref{Fig3} (used also for the RL models) with the exception of the removal of the LSTM layer which is not needed since we aim to decode from the initial configuration of each episode. The model was trained by supervised learning with a cross-entropy loss function to predict whether food is visible from the dominant agent's point of view. 

Fig \ref{Fig4} shows the accuracy at validation samples as a function of the number of training epochs. As observed in Fig \ref{Fig4} when visual input is fed in allocentric coordinates the model exhibits a quicker learning of the decision of whether the food item is observed by the dominant agent. In particular, with the allocentric mode of visual processing the decision is learn with high accuracy ($>90\%$) from 4 epochs of training. Similar level cannot be achieved by egocentric viewpoint in 20 epochs of training (reaching $\sim 83\%$), an amount of training by which allocentric view already provides close to perfect decoding. 

\begin{figure}[!ht]
%\hspace{-5.75cm}
%\begin{minipage}[r]{1.43\textwidth}
 \includegraphics[width=\linewidth]{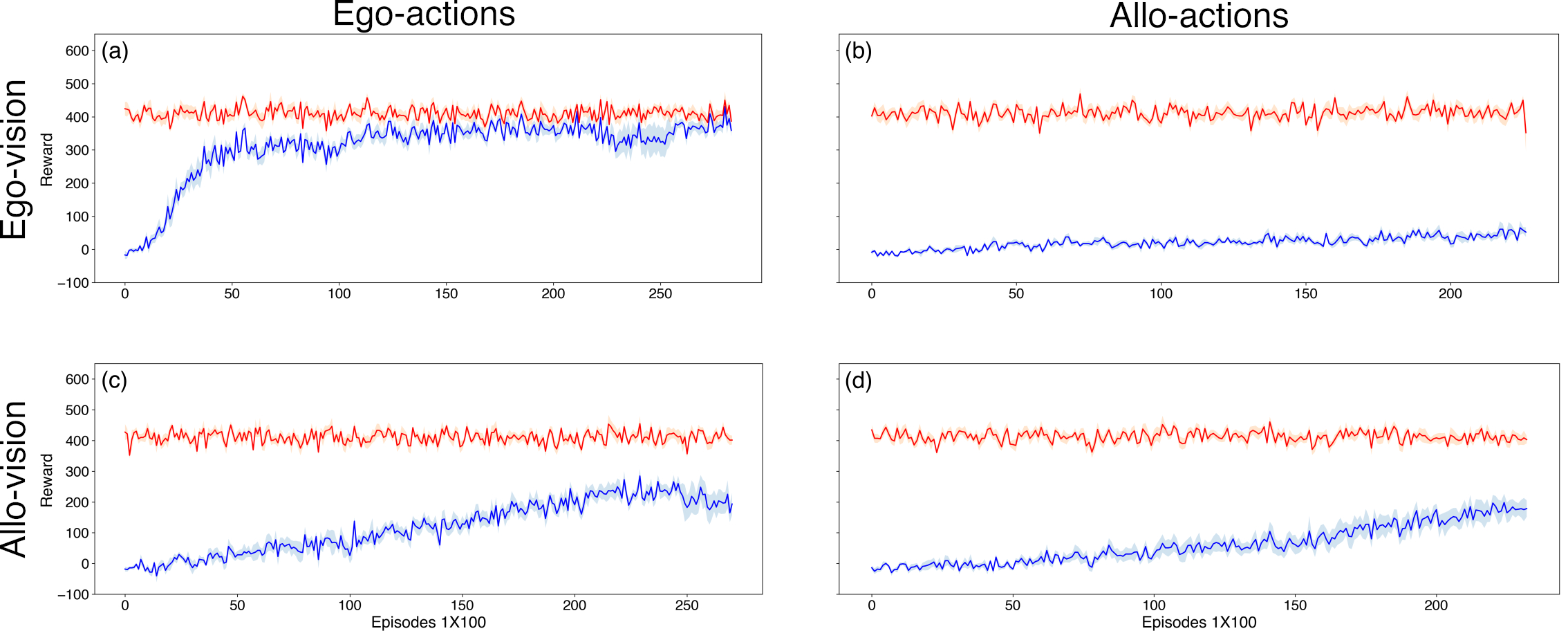}
\caption{Average reward obtained by the \textit{Subordinate} agent (blue) over 100 episodes per data point. The red line shows the maximal reward that would be possible to obtain over the same episodes.  Results were averaged over 7 different seeds. Upper and bottom rows illustrate the reward during egocentric and allocentric vision, respectively. Left and right columns illustrate the reward during using egocentric and allocentric encoding of actions, respectively. Fluctuations in the maximal reward (red line) are due to the different fraction of episodes (within each set of 100) in which an optimal agent should go or avoid the food and the very different reward for the correct decision in each case.}
\label{Fig5}
%\end{minipage}
\end{figure}

We then proceed to compare how an egocentric and allocentric encoding of visual input affects the performance of an agent in solving the perspective taking task when learning from trial and error. We note that the full task now involves the need of not only integrating the necessary information to determine whether to approach the food, but also to maintain and convert this information into a motor plan. For this study we also considered the egocentric and allocentric representations for moving actions in addition to the egocentric and allocentric encodings for visual perception. Thus, we explore 4 combinations of visual and action representations (ego\textsubscript{vis}-ego\textsubscript{act}, allo\textsubscript{vis}-ego\textsubscript{act}, ego\textsubscript{vis}-allo\textsubscript{act}, and  allo\textsubscript{vis}-allo\textsubscript{act}).

In this case agents who had egocentric visual input and action encoding quickly learned the task, whereas agents with allocentric input were not able to efficiently learn the perspective taking task irrespective of the action encoding (see Fig \ref{Fig5}).

%To better understand why this is the case we trained four different agents with two different types of input (egocentric and allocentric) and two types of action (egocentric and allocentric) and assessed their ability to learn the perspective taking task.

%\begin{equation}
%    NextAction_{e,e} = F(goal_{pos})
%\end{equation}
%\begin{equation}
%    NextAction_{e,a} = F(goal_{pos},Subordinate_{orientation})
%\end{equation}
%\begin{equation}
%    NextAction_{a,e} = F(goal_{pos},Subordinate_{orientation})
%\end{equation}
%\begin{equation}
%    NextAction_{a,a} = F(goal_{pos})
%\end{equation}

Indeed, we can see from Fig \ref{Fig5} that out of the four combinations, only the "ego\textsubscript{vis}-ego\textsubscript{act}" condition performed well in the task. This is at first surprising, given that: i) the previous result that without navigation (in the supervised learning setting) the allocentric agents are better in deciding whether the dominant can see the food, and ii) the agents with allocentric vision underperform the agents with egocentric vision even when using the same action encoding. Also, agents with egocentric vision cannot learn the task when using allocentric actions compared to the efficient and almost perfect score when using also egocentric actions.

Hence, part of the difficulty seems to come from the coupling of visual to navigational aspects, i.e. not only extracting the relevant information from the visual input but also its conversion into the appropriate actions. No single factor (visual encoding or action encoding) seems to individually explain the success at efficiently learning the task, rather what matters is their specific combination. Therefore, we analysed in more detail the required computations for the different cases to succeed in solving the task. 

In the case of agents with allocentric actions (right column in Fig. \ref{Fig5}) the movements are in reference to fixed directions in the space, namely North, South, West, East, and standstill. This implies that when the \textit{Subordinate} agent with egocentric vision but allocentric actions (ego\textsubscript{vis}-allo\textsubscript{act}) sees food in front of it, it will not automatically know which action it needs to take to get closer to the food. In comparison to that, with egocentric actions (ego\textsubscript{vis}-ego\textsubscript{act}) it is enough to move forward to approach the food. Similarly, simple heuristics also exist whenever the food item is within the field of vision of the agent. Thus, when working with egocentric actions, for agents with egocentric vision the selection of the optimal action is solely a function its location relative to the goal, an information which is directly accessible from its visual encoding. However, when working with allocentric actions, the same agent would need to know their own orientation which is not directly accessible with an egocentric visual encoding. Hence, the difference of performance between the panels a) and b) in Fig. \ref{Fig5} when learning by trial and error.

With allocentric vision (for which the learning of whether the food is visible by the dominant was best in the supervised setting), the performance and speed of learning by reinforcement are significantly inferior to the case with egocentric vision. In this case, agents with allocentric visual input also need to take into account their orientation in order to choose the appropriate action towards the goal. For example, if the agent uses an allocentric viewpoint but an egocentric action space, some of the simple heuristics available for egocentric representations such as "move forward when food is in front of you" seem harder to discover. This might be due to the fact that additional processing is required to extract the relevant variables (e.g. is the food item in front of the agent?) to implement certain heuristics using allocentric visual encoding.

So far, we have quantified the success of the different types of \textit{Subordinate} agents by the amount of reward obtained. Next, we refine the analysis of the success by evaluating which behavior and types of errors the agents committed. In particular, we tested all 4 models by initializing the agents and environment in all possible allowed combinations. Then we counted in how many episodes the agent performed a correct or incorrect behavior keeping track on whether in the given trial the food item should have been eaten or not (food item is within the visual field of the dominant agent). 

\begin{figure}[!ht]
%\hspace{-2cm}
%\begin{minipage}[r]{1.1\textwidth}
\centering
 \includegraphics[scale=0.91]{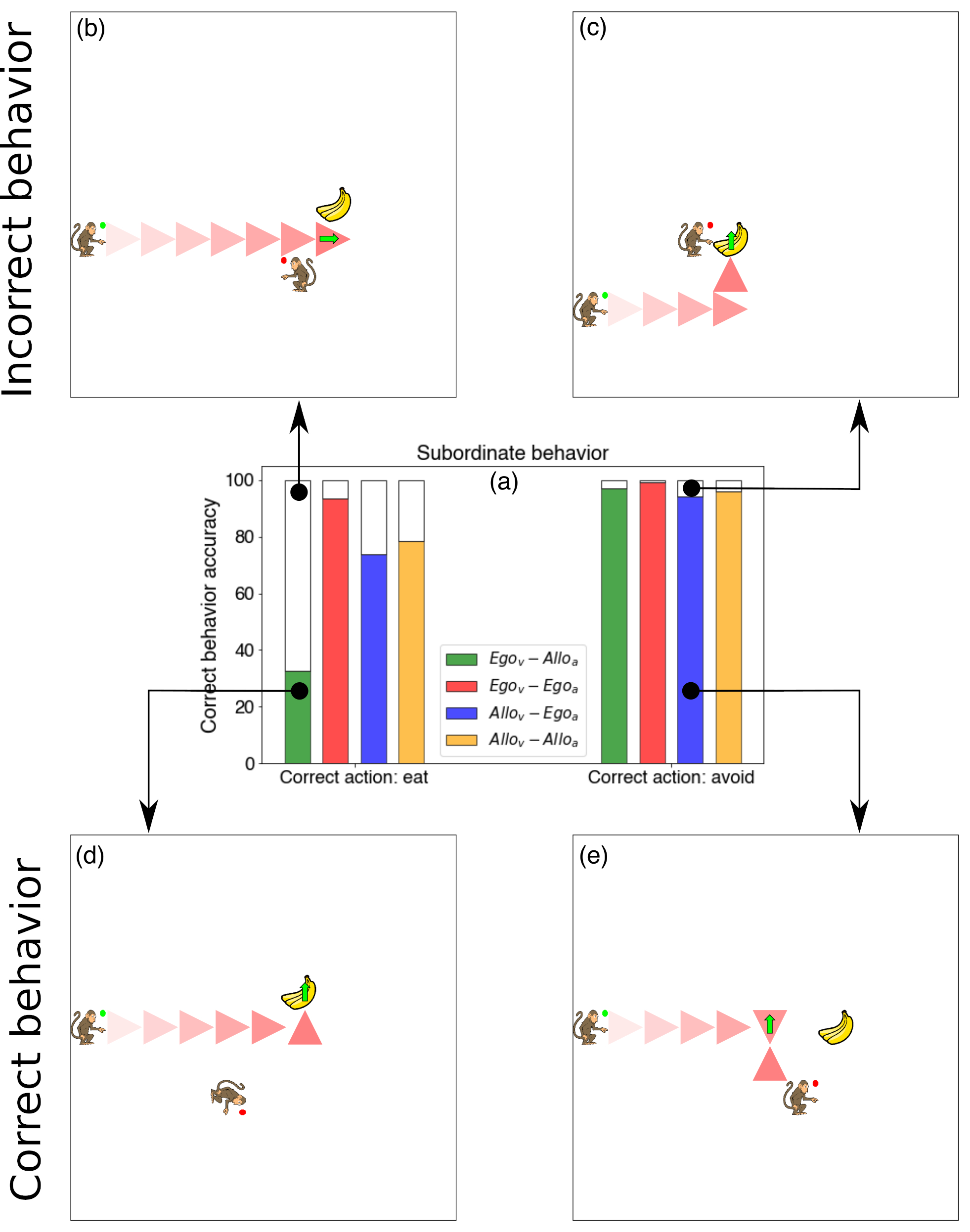}
\caption{Quantification of the subordinate behavior and examples of model trajectories. \textbf{a)} Bar plot with the percentage of correct behavior depending on the type of trial (eating food is rewarded vs penalised). \textbf{b)} Example of the subordinate agent (green circle) avoiding the food although it should approach it. \textbf{c)} Example of the model reaching the food when it should not reach it. \textbf{d)} Example of the model performing the correct behavior of navigating and obtaining the food. \textbf{e)} Example of model behavior of avoiding the food when this is observed by the dominant agent (red circle).}
\label{Fig6}
%\end{minipage
\end{figure}

Fig \ref{Fig6}a shows the results of the analysis. The agent equipped with egocentric viewpoint and action space (ego\textsubscript{vis}-ego\textsubscript{act}), obtained the food item in more than $93.38\%$ of the cases when it should eat the food (non-observed by the dominant) and avoided it $99.3\%$ of the cases when it should avoid it (observed by the dominant). Compared to the $78.49\%$ and $95.89\%$ for the allo\textsubscript{vis}-allo\textsubscript{act} case, we see that this type of agent had a main issue in obtaining the food when this was edible (not seen by the dominant). Similar performance in both types of trials is realized by the agent endowed with egocentric actions (allo\textsubscript{vis}-ego\textsubscript{act}). Most dramatically, agents with allocentric visual processing and action space obtain the food item in only $32.59\%$ of the cases when the food is rewarded.

\begin{figure}[!ht]
%\hspace{-5.75cm}
%\begin{minipage}[r]{1.43\textwidth}
\includegraphics[width=\linewidth]{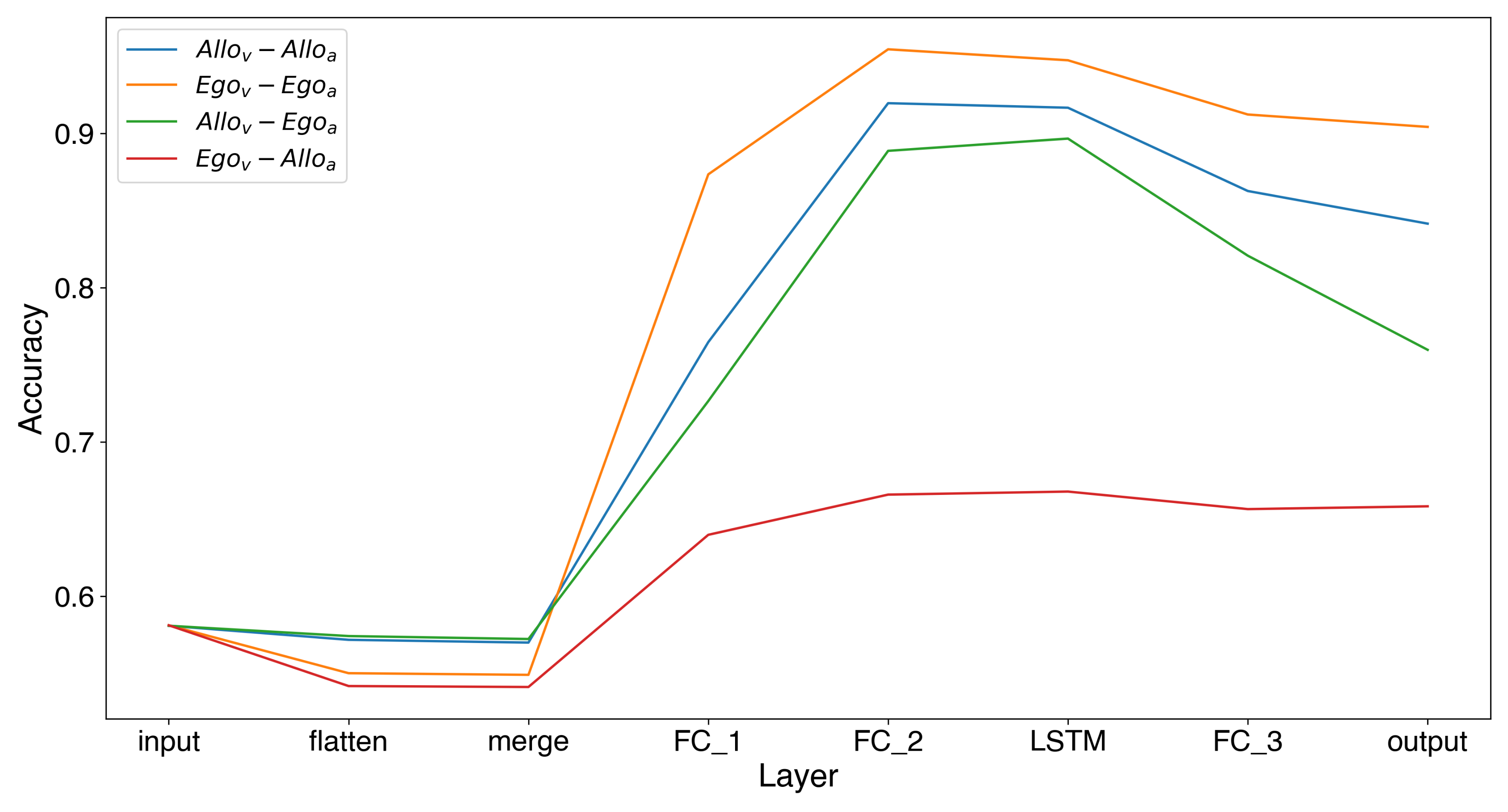}
\caption{Accuracy of a linear probe in predicting if \textit{Dominant} sees the food or not from different layers of the architecture. The layers studied are Input (raw input), flatten (after convolution without orientations), merge (with orientations), FC\_1 (First fully connected layer), FC\_2, LSTM, FC\_3 (fully connected layer), and output (the actions Q values).}
\label{Fig7}
%\end{minipage}
\end{figure}

Finally, we also analysed the information that each layer of the architecture contains about whether the food is observed by the dominant agent or not. This is the essential bit of information that needs to be extracted from the visual input to guide the decision of the agent of whether to approach the food or not. To this end we added a linear decoder in each layer of the neural architecture for each of the 4 RL models. That is, we trained a linear decoder (linear discriminant analysis) from different layers of the architecture to predict whether the food is visible by the dominant agent. Fig \ref{Fig7} shows that indeed it is possible to reliably decode this information from 3 out of the 4 types of trained agents. Also we observed that the information about whether the food is visible from the dominant agent increases with further layers of processing until peaking around the last convolutional layer and the LSTM layer to slightly decrease near the output layer.

%We used the best seed of each experiment.(allo-allo =allocentric vision - allocentric actions, etc...).

\section{Discussion}
In this work we aimed to develop RL agents that could solve a basic perspective taking task. For that we devised a perspective taking task inspired by work done with chimpanzees. Our goal was not to design a comprehensive model of perspective taking behavior in humans or other animals. Rather, we explored whether relatively simple deep reinforcement learning algorithms can capture basic computational aspects of perspective taking. The behavior of the agents showed evidence for basic perspective taking skills, which shows that at least a part of perspective taking skills might indeed be learned through RL. 

The real advantage of AI algorithms is that they allow us to deconstruct the studied process. In this case we deconstructed the perspective taking task by a) separating the decision and navigational aspects of the task and b) using different combinations of visual input (ego-vision vs allo-vision) and action (ego-action vs allo-action).

\subsection{Allocentric vs egocentric perspective taking}
The condition with egocentric vision corresponds to the natural way how animals interact with the world: they perceive objects and other agents from their viewpoint. However, for some reason the brains of animals have also developed specific systems where objects are represented in allocentric fashion - they are represented in relation to each other as in a map \cite{burgess2001memory, chersi2015cognitive, wang2020egocentric}. This allocentric representation allows animals to compute certain aspects of the world more easily. One of such functions might be perspective taking. Indeed, in our work we found that in the supervised setting without navigation agents can much more readily learn perspective taking skills from allocentric input representations. Had we stopped there we would have concluded that the allocentric input representation is the optimal one for solving tasks involving perspective taking.

However, we then studied agents in the RL setting, where the agents needed to demonstrate their perspective taking skills by navigating toward the food item. Under these settings we found that in the simple environment studied ego\textsubscript{vis}-ego\textsubscript{act} agents clearly outperformed all other visual-action combinations, including the allo\textsubscript{vis}-allo\textsubscript{act} case. This seems to show that when navigation is involved the egocentric representation is actually much more efficient for learning. That said, it has to be kept in mind that we studied a very simple setting. In all of our experiments, the goal was in the field of the vision of the \textit{Subordinate}. Hence, this is like navigating toward a building you see in front of you. Using a map probably will make things more complicated. 

An important question in neuroscience is how this transformation from egocentric to allocentric coordinates is computed in the brain \cite{chersi2015cognitive, bicanski2018neural}. Also, it is clear that in the animals these two systems interact \cite{wilber2014interaction, bicanski2018neural, wang2020egocentric}. In the present work we did not study the interactions of these two systems. In the future work we seek to study how the allocentric representation is computed from the egocentric input and how these two systems interact during online decision making. 

\subsection{Different levels of perspective taking in humans and AI}

Although some agents in this study could solve the visual perspective taking tasks presented here, we do not claim that the complexity of the task is anywhere near what humans encounter in their everyday perspective taking tasks \cite{apperly2011mindreaders}. Also, although the current task was inspired by work done with chimpanzees, the chimpanzees were not trained on the task, they were just tested at it- they had acquired perspective taking from various encounters with different chimpanzees under natural conditions \cite{hare2000chimpanzees}. Furthermore, our RL agents could compute their decision to approach or avoid the food based on the orientation of the dominant and the food position, whereas in the original experiments with chimpanzees \cite{hare2000chimpanzees} the different conditions also involved obstacles (to hide the food). Hence, our current setup is a very simplified version of perspective taking, but hopefully it lies the groundwork for more elaborate experiments. 

Based on human studies, perspective taking has been divided into two levels \cite{apperly2011mindreaders}, namely level 1 and 2 perspective taking. Level 1 perspective taking is about the question "what the other agent sees" and level 2 is more complicated, by also taking into account how it is seen \cite{apperly2011mindreaders}. Our agents mastered level 1 perspective taking, but even here the caveat is that our RL agents were trained for thousands of episodes in the very same task. When children and chimpanzees are given level 1 perspective taking tasks, they solve it without training \cite{apperly2011mindreaders}. In this sense we advocate for claiming that the current RL agents solve level 0 perspective taking tasks that we define as "achieving perspective taking level 1 behavior after extensive training". From this viewpoint it is also clear which tasks should our RL agents try to solve next - level 1 and 2 perspective taking without extensive training on the same task. 
Rabinowitz and colleagues \cite{rabinowitz2018machine} used a meta-learning strategy to close the gap between level 0 and level 2 perspective taking, however their study had several assumptions (training by supervision, full observability and non-behaving agents) that were addressed in our study. In general, a future avenue of work should address more complex environments and perspective taking tasks involving more agents and stringent generalization tests.

Another avenue for future work is that of opening the networks that successfully implement perspective-taking capabilities during RL. In particular, it will be interesting to search and study the receptive fields of specific neurons in the network whose activity correlates with a decision requiring perspective-taking skills. 

\section{Conclusion}
Perspective taking, like any other cognitive ability, has multiple facets and for the scientific understanding of such abilities it is necessary to study all of these facets \cite{apperly2011mindreaders}. Here we studied the simplest possible case where agents controlled by artificial neural networks learned with the help of reinforcement learning in a simple task. 

Investigating the capabilities and limitations of RL agents in acquiring perspective taking is a first step towards dissecting the algorithmic and representational options underlying perspective taking and more generally, Theory of Mind. As human communication heavily relies on Theory of Mind \cite{tomasello2010origins, tomasello2019becoming, scott2014speaking}, a better understanding of Theory of Mind is a prerequisite for developing AI algorithms that can take the perspective and comprehend the intentions of humans.

\section{Acknowledgments}
We would like to thank Daniel Majoral for fruitful discussions and comments.

\bibliographystyle{unsrt}  
\bibliography{references}  %%% Remove comment to use the external .bib file (using bibtex).
\end{document}